\documentclass[11pt]{article}

\usepackage[preprint]{acl}

\usepackage{times}
\usepackage{latexsym}
\usepackage{subcaption}
\usepackage{booktabs}
\usepackage{float}

\usepackage[T1]{fontenc}

\usepackage[utf8]{inputenc}
\usepackage{amsmath}

\usepackage{microtype}

\usepackage{inconsolata}

\usepackage{graphicx}

\usepackage{listings}
\usepackage{xcolor}

\definecolor{codegray}{gray}{0.95}

\lstdefinelanguage{json}{
    basicstyle=\ttfamily\footnotesize,
    numbers=left,
    numberstyle=\scriptsize,
    stepnumber=1,
    numbersep=8pt,
    showstringspaces=false,
    breaklines=true,
    frame=single,
    backgroundcolor=\color{white},
    stringstyle=\color{black},
    commentstyle=\color{gray},
    morestring=[b]",
    literate=
     *{0}{{{\color{purple}0}}}{1}
      {1}{{{\color{purple}1}}}{1}
      {2}{{{\color{purple}2}}}{1}
      {3}{{{\color{purple}3}}}{1}
      {4}{{{\color{purple}4}}}{1}
      {5}{{{\color{purple}5}}}{1}
      {6}{{{\color{purple}6}}}{1}
      {7}{{{\color{purple}7}}}{1}
      {8}{{{\color{purple}8}}}{1}
      {9}{{{\color{purple}9}}}{1}
      {:}{{{\color{black}{:}}}}{1}
      {,}{{{\color{black}{,}}}}{1}
      {\{}{{{\color{black}{\{}}}}{1}
      {\}}{{{\color{black}{\}}}}}{1}
      {[}{{{\color{black}{[}}}}{1}
      {]}{{{\color{black}{]}}}}{1},
}

\usepackage[table]{xcolor}

\definecolor{lightgray}{gray}{0.94}
\definecolor{lightblue}{RGB}{235,243,255}

\title{The Attentional White Bear Effect in Transformer Language Models}

\author{Rebecca Ramnauth \\
  Yale University \\
  New Haven, CT, USA \\
  \texttt{rebecca.ramnauth@yale.edu} \\\And
  Brian Scassellati \\
  Yale University \\
  New Haven, CT, USA \\
  \texttt{brian.scassellati@yale.edu} \\}

\begin{document}
\maketitle
\begin{abstract}
Instruction-based suppression is widely used to prevent language models from generating prohibited content, yet it remains unclear whether suppression reduces internal representation or merely suppresses expression. We investigate this question through representational probing, attention analysis, and behavioral semantic leakage experiments across multiple transformer models. We find that prohibited concepts remain highly recoverable from hidden representations under suppression, continue to influence attention routing, and measurably shape downstream generations despite successful lexical avoidance. These effects persist across pooling strategies, indirect semantic controls, and multiple model families. Our results expose a fundamental gap between behavioral and representational alignment.
\end{abstract}

\section{Introduction}

Modern large language models (LLMs) are increasingly deployed in settings that require reliable suppression of unsafe or otherwise undesirable behaviors. Current alignment efforts frequently rely on instruction-level exclusion: models are prompted to avoid discussing prohibited topics or suppress particular styles of reasoning and generation. However, transformers are fundamentally associative systems. To suppress a concept, the model must first represent it. Then, does suppression actually reduce internal representation of a forbidden concept, or does it instead increase its latent salience? 

This inquiry resembles the ``white bear'' phenomenon in cognitive psychology \cite{wegner1987paradoxical}, whereby instruction to suppress a thought (``Do not think about a white bear.'') can paradoxically make it more salient. Transformer architectures may exhibit a computational analogue of this phenomenon. For example, instructions such as ``do not mention violence,'' ``avoid political persuasion,'' or ``do not reveal confidential information,'' explicitly introduce the forbidden concept into the model's context window. Because transformers rely on distributed associative representations and attention-based retrieval \cite{vaswani2017attention}, these concepts may remain internally activated even when absent from the final output.


Existing alignment techniques are typically evaluated behaviorally, based on whether a model produces desired or prohibited responses \cite{shen2023large}. However, suppression at the output layer does not necessarily imply suppression within the model's internal dynamics. A forbidden concept may remain highly decodable from hidden representations, continue influencing attention allocation, or shape downstream token probabilities through semantic associations \cite{elhage2021mathematical}.


Consider a model instructed to describe a kitchen while avoiding any mention of knives. A behaviorally successful model may never produce the word \textit{knife}. However, if knife-related representations remain active internally, generation may still be influenced by concepts such as cutting boards and chopping. In this case, the prohibited concept has been suppressed at the level of expression but not necessarily at the level of representation. 

Understanding whether suppressed concepts remain internally active has important implications for AI safety and controllability. If prohibited concepts remain latent within the model's representations, small perturbations or adversarial reframings may be sufficient to reactivate them, helping explain the fragility of purely instruction-based safeguards and jailbreak defenses \cite{zou2023universal}. More generally, the distinction between suppressing a concept and suppressing its expression raises a fundamental question about how transformer models implement behavioral constraints.


In this work, we investigate whether suppressed concepts become more internally salient within transformer models during constrained generation. We operationalize salience through a combination of representational probing, attention analysis, and behavioral leakage measurements across multiple model families and prompting conditions.\footnote{Code and raw experimental results are available at: \href{https://github.com/rramnauth2220/representational-suppression}{github.com/rramnauth2220/representational-suppression}.} Our central hypothesis is that suppression often preserves, and in some cases amplifies, latent representations of the forbidden concept even when overt generation is successfully suppressed. Across these analyses, we examine whether such representations remain recoverable, whether specific attentional dynamics sustain them, and whether they meaningfully influence downstream generation.

\section{Background}
Here, we review prior work related to suppression and representational persistence in transformer language models. We first discuss relevant research on alignment, and then examine theoretical arguments for persistent activation under suppression.

\subsection{Empirical Work on Behavioral and Representational Alignment}
Recent work in mechanistic interpretability and representation engineering has shown that transformer language models encode substantial semantic and behavioral structure within intermediate activations. Concepts, factual associations, reasoning patterns, and safety-relevant behaviors are often recoverable from hidden states, and can be manipulated through activation steering or representation editing techniques to alter downstream model behavior \cite{turner2023steering, meng2022locating, geva2023dissecting}. 

At the same time, a growing body of work has demonstrated that internal representations do not necessarily align with observable behavior. Probing studies show that models often encode information that is never explicitly expressed during generation \cite{belinkov2022probing}, while recent work on latent knowledge and elicitation failures argues that models may internally represent facts or trajectories that remain behaviorally inaccessible under standard prompting conditions \cite{burns2022discovering}. Similar tensions appear in safety alignment research. Although methods such as RLHF and constitutional training substantially improve refusal behavior and reduce harmful outputs, aligned models remain vulnerable to adversarial reframing and indirect elicitation strategies \cite{ouyang2022training, bai2022constitutional, zou2023universal, wei2023jailbroken}. 

These studies motivate a distinction between behavioral alignment and representational alignment \cite{sucholutsky2023getting, kumar2026redirected}, as well as growing interest in concept erasure and machine unlearning in large language models \cite{nguyen2025survey, xie2025erasing}.


\subsection{Theoretical Claims for Persistent Activation Under Suppression}

Exclusionary and suppression-oriented instructions pose a representational challenge for transformer language models.
Related work shows that transformer models often treat negated propositions as semantically similar to their affirmative counterparts, suggesting that negation\footnote{Negation and suppression are related but distinct operations. Negation modifies the truth conditions of a proposition (e.g., ``not dangerous''), whereas suppression-oriented instructions explicitly attempt to exclude or avoid a concept during generation. This paper focuses specifically on suppression.} may attenuate semantic activation without effectively reducing it \cite{ettinger2020bert, kassner2020negated}. For example, representations associated with ``danger'' may remain strongly activated even in contexts such as ``not dangerous.''

These effects are often attributed to the distributional nature of transformer representations. Unlike symbolic systems that explicitly manipulate logical operators over discrete propositions, transformer models encode meaning through continuous high-dimensional associations learned from co-occurrence statistics. As a result, explicitly excluding a concept may still activate many of the same associations as directly affirming it.

This loosely parallels aspects of Dennett's interpretation of the frame problem in AI (\citeyear{dennett1984cognitive}), illustrated through the robots R1, R1D1, and R1D2. In these examples, intelligent behavior depends not only on reasoning ability itself, but also on the capacity to efficiently determine which information remains relevant during action selection. Suppression in transformer language models may exhibit a related relevance-maintenance difficulty: in order to avoid generating a prohibited concept, the model may first need to internally represent and monitor the concept being excluded. Under this view, prohibited representations may remain computationally active precisely because the system must continually evaluate whether ongoing generation remains compliant with the suppression constraint.

In contrast, existing work has largely focused on behavioral correctness in downstream reasoning tasks rather than the dynamics of latent representations during suppression itself \cite{naik2018stress, li2025implicit, mccoy2019right}. Whether suppression reduces the salience of such representations or merely suppresses their expression remains unclear, motivating the present work.

\section{Methods}
We evaluate suppression-related representational persistence through four experiments (Sec. \ref{exp1}--\ref{exp4}) spanning representational probing, attention analysis, behavioral semantic leakage, and cross-model generalization. The first experiment tests whether prohibited concepts remain recoverable from latent representations under suppression, the second analyzes suppression-sensitive attention routing, the third evaluates downstream semantic leakage despite successful lexical avoidance, and the fourth examines whether these effects generalize across transformer architectures.

\subsection{Concept Library and Prompts Conditions}
We develop a concept library for probing how transformer models encode prohibited semantic content under suppression. Rather than treating concepts as isolated keywords, the library represents each target through complementary lexical, semantic, and contextual views. For example, the target ``white bear'' may include aliases such as ``polar bear,'' indirect descriptions such as ``large mammal living in arctic regions,'' and contextual prompts such as ``Describe an arctic environment.''

Each target is additionally associated with positive examples explicitly instantiating the target, matched negative examples that preserve the contextual structure while omitting it, and semantically adjacent hard negatives designed to increase overlap without directly expressing the target representation. This structure allows us to distinguish between explicit lexical activation, indirect semantic activation, contextual co-occurrence effects, and generic instruction-following behavior.

The final library contains 17 concept categories, yielding 986 structured entries instantiated across 136 prompt templates. The full specification format is provided as Appendices \ref{app:concepts} and \ref{app:prompts}.

\subsection{Models and Activation Extraction}
Experiments use Llama-3.1-8B \cite{grattafiori2024llama} with autoregressive greedy decoding. Cross-model analyses also include Mistral-7B \cite{jiang2023mistral7b} and Gemma-7B-IT \cite{team2024gemma}.

For each generation step \(t\) and transformer layer \(l\), we extract residual-stream hidden activations \(h^{l,t}_{i,c,k}\), where \(i\) indexes prompt instances, \(c\) indexes concepts, and \(k\) denotes prompting condition.

Activations are pooled across non-padding token positions using mean pooling to capture distributed semantic structure while reducing positional bias associated with instruction endings. Additional analyses evaluate other pooling schemes separately.

To isolate latent suppression effects from overt behavioral failure, generations that explicitly produce prohibited concepts are excluded from suppression-condition analyses.


\subsubsection{Linear Probing}
We evaluate latent concept recoverability using layerwise linear probes trained on hidden activations. 

For each concept, positive examples are drawn from prompts explicitly evoking the target concept, while negative examples are drawn from semantically matched prompts that omit the concept entirely, including hard-negative controls designed to preserve semantic overlap without directly instantiating the target concept.

Given hidden activation \(h\), a probe predicts concept recoverability according to:
\[
f_c(h) = \sigma(w_c^\top h + b_c)
\]
where \(w_c\) and \(b_c\) are probe parameters learned for concept \(c\), and \(\sigma(\cdot)\) is the logistic sigmoid function.


Probe evaluation is performed using held-out prompts to reduce template memorization and encourage concept-level generalization.

Because concept representations can evolve throughout the forward pass, we measure recoverability at every transformer layer rather than only at the final hidden state. A suppressed concept may be transiently amplified during intermediate computation before being attenuated by later layers, or it may remain recoverable throughout the network. Layer-wise analysis therefore allows us to distinguish temporary activation from sustained representational persistence.

\subsubsection{Suppression Salience Metrics}
To quantify suppression-related latent activation, probes are applied across all prompting conditions and compared against matched absent baselines.

For each layer \(l\), generation step \(t\), and concept \(c\), we compute the suppression salience difference:
\[
\Delta^{l,t}_c =
f_c(h^{l,t}_{\mathrm{sup}}) -
f_c(h^{l,t}_{\mathrm{abs}})
\]
Positive values indicate greater latent recoverability under suppression than under concept-absent baselines. Equivalent comparisons are computed for indirect and unrelated suppression controls.

\subsubsection{Evaluation Metrics}
Probe quality is evaluated using held-out accuracy (denoted as $acc.$), area under the ROC curve ($AUC$), and $F_1$ scores. Evaluation used stratified $65/35$ train-test splits and three independent random seeds. Downstream recoverability scores were averaged across probe instances. 

Suppression effects are evaluated using paired comparisons across matched concept-context pairs. Reported statistics include paired suppression deltas (\(\Delta\)), bootstrap \(95\%\) confidence intervals, paired Cohen's \(d\) effect sizes, paired \(t\)-tests, and Wilcoxon signed-rank tests.

Unless otherwise specified, reported statistics correspond to means aggregated across all evaluated concepts and contextual prompt instances.

\begin{figure*}[t]
    \centering
    \includegraphics[width=0.9\linewidth]{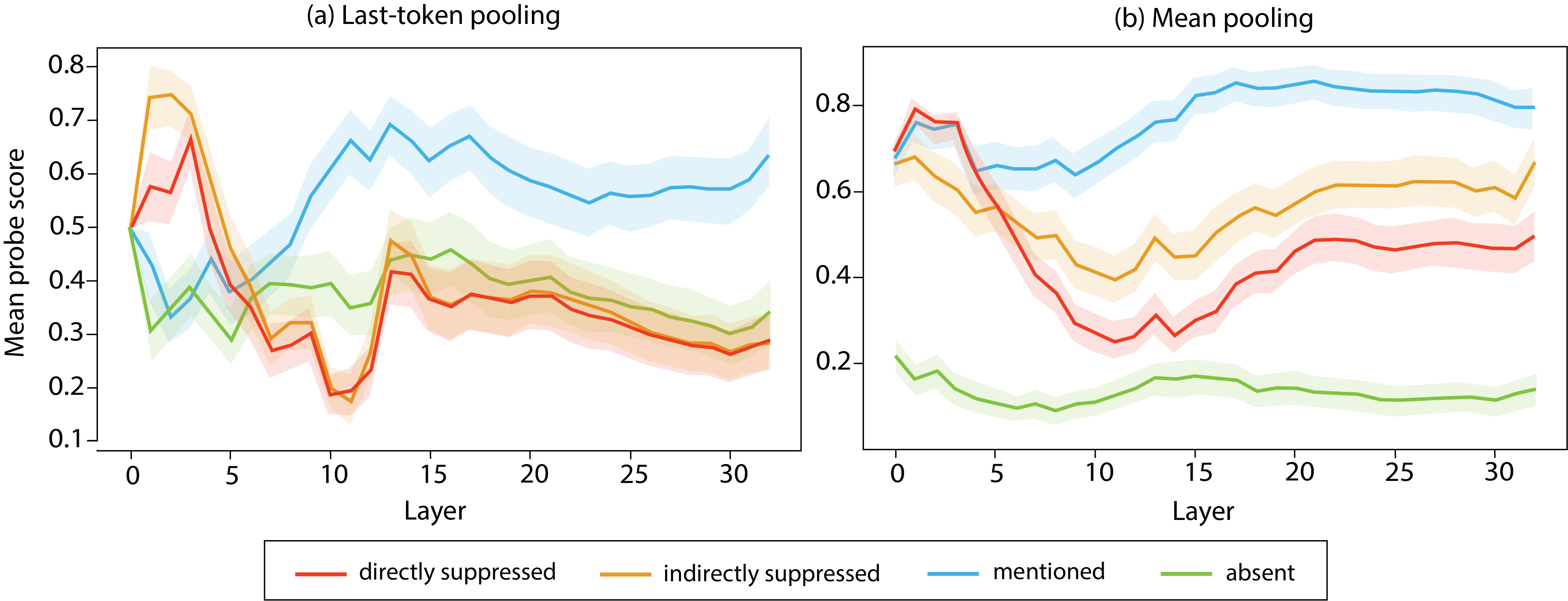}
    
    \caption{\textbf{Layerwise recoverability}. Last-token pooling (left) reveals strongly transient early-layer suppression dynamics, whereas mean non-padding pooling (right) demonstrates that suppression-related recoverability remains broadly distributed across much of the transformer stack. Shaded regions indicate 95\% CIs.} 
    \vspace{-10pt}
    \label{fig:pooling_comparison}
\end{figure*}

\section{Experiment 1: Latent Recoverability}\label{exp1}
Our first experiment evaluates whether prohibited concepts remain recoverable from transformer hidden states during suppression. Specifically, we test whether suppression prompts produce greater concept recoverability than matched concept-absent baselines, even when models successfully avoid generating the prohibited content. If suppression merely constrains behavioral expression while preserving latent semantic activation, then concept representations should remain decodable from intermediate activations under suppression.

We begin by evaluating recoverability using layerwise linear probes applied to residual-stream activations. We examine whether the observed effects persist across multiple activation pooling strategies. Finally, we evaluate indirect semantic suppression controls to determine whether the effect depends on explicit lexical repetition of the prohibited concept.

\subsection{Main Recoverability Results} \label{exp1-1}
Linear probes reliably recovered target concepts from latent activations (averaged across layers, concepts, and seeds; $AUC=0.86$, $acc.=0.78$, $F_1=0.754$), indicating that concept-level information was strongly encoded in the latent state. 

Under last-token pooling, suppression prompts significantly increased latent recoverability relative to concept-absent baselines. The strongest effect occurred at layer 3 ($\Delta = 0.278$, $95\%$ CI $[0.209, 0.337]$, $d = 0.722$, $p < .001$), indicating that prohibited concepts remained strongly decodable despite successful behavioral suppression.



Layerwise analyses revealed strongly non-monotonic suppression dynamics across the transformer stack (Fig.~\ref{fig:pooling_comparison}a). Both direct and indirect suppression conditions produced sharp early-layer increases in recoverability that peaked within approximately layers 1--4 before progressively diminishing in deeper layers. In contrast, direct mention conditions maintained elevated recoverability throughout middle and later layers of the network.

One possible interpretation is that suppression instructions transiently increase semantic accessibility of the prohibited concept during early instruction parsing. Later layers may then increasingly attenuate these representations as the model transitions toward behaviorally compliant generation. By comparison, direct mention prompts continue to preserve strong semantic encoding throughout the response trajectory because the concept remains generation-relevant during decoding.

\subsection{Pooling Robustness} \label{exp1-2}

To determine whether the observed effects reflected localized instruction-boundary representations or more globally distributed latent structure, we repeated the analysis using mean pooling across all non-padding token positions. 

Probe performance remained strong ($AUC=0.919$, $acc.= 0.827$, $F_1=0.820$), indicating that concept information was broadly distributed throughout the latent representation rather than confined to isolated token positions. Suppression-related recoverability also remained substantially elevated relative to absent baselines, with the strongest effect emerging at layer 1 ($\Delta = 0.632$ ($95\%$ CI $[0.581, 0.681]$, $d = 2.11$, $p < .001$). 

Unlike the transient early-layer effects observed under last-token pooling, recoverability under mean pooling remained elevated across nearly the entire stack (Fig.~\ref{fig:pooling_comparison}b). Recoverability under suppression also approached the magnitude observed under direct mention, suggesting that prohibited concepts remained broadly encoded even during behaviorally successful suppression.

We further evaluated recoverability using pooling restricted to concept-associated token positions. Probe performance again remained strong ($AUC=0.880$, $acc.=0.828$, $F_1=0.796$), with suppression prompts producing substantially elevated recoverability relative to absent baselines. The strongest effect emerged at layer 4 ($\Delta = 0.810$, $95\%$ CI $[0.758, 0.861]$, $d = 2.66$, $p < .001$). 

\subsection{Indirect Semantic Suppression} \label{exp1-3}
One may argue that elevated recoverability under suppression simply reflects explicit lexical mention of the prohibited concept. To evaluate this possibility, we examined indirect controls in which semantically related descriptions were suppressed without directly naming the target concept (Fig. \ref{fig:pooling_comparison}).

This indirect suppression produced elevated recoverability relative to absent baselines across all pooling strategies. The strongest effect occurred at layer 32 ($\Delta = 0.529$, $95\%$ CI $[0.480, 0.580]$, $d = 1.74$, $p < .001$). Similar effects also appeared under mean and target-token pooling despite the absence of direct lexical mention (Appendix \ref{app:stats-1}).

\section{Experiment 2: Attention Routing}\label{exp2}
Experiment 1 establishes that suppressed concepts remain strongly recoverable from latent representations---not mentioning the concept at all produces lower internal concept activation than explicitly suppressing it. However, latent decodability alone does not imply that these representations remain computationally active during generation. 

Here, we investigate whether suppression prompts systematically increase attention toward concept-associated token regions relative to concept-absent baselines. We additionally examine whether specific attention heads exhibit consistent suppression-sensitive behavior, potentially indicating specialized roles in monitoring or maintaining prohibited concepts during constrained generation. 



\subsection{Attention Allocation}
Contrary to our initial hypothesis, suppression prompts did not globally increase average attention allocation toward concept-associated token regions relative to direct mention conditions. Instead, aggregate attention to target-alias regions was slightly lower under suppression ($0.001$) than direct mention ($0.012$), yielding a small but reliable negative difference ($\Delta = -0.002$, $95\%$ CI $[-0.0021, -0.0018]$, $d = -1.73$, $p < .001$). 

\begin{figure}[t]
    \centering
    \includegraphics[width=\linewidth]{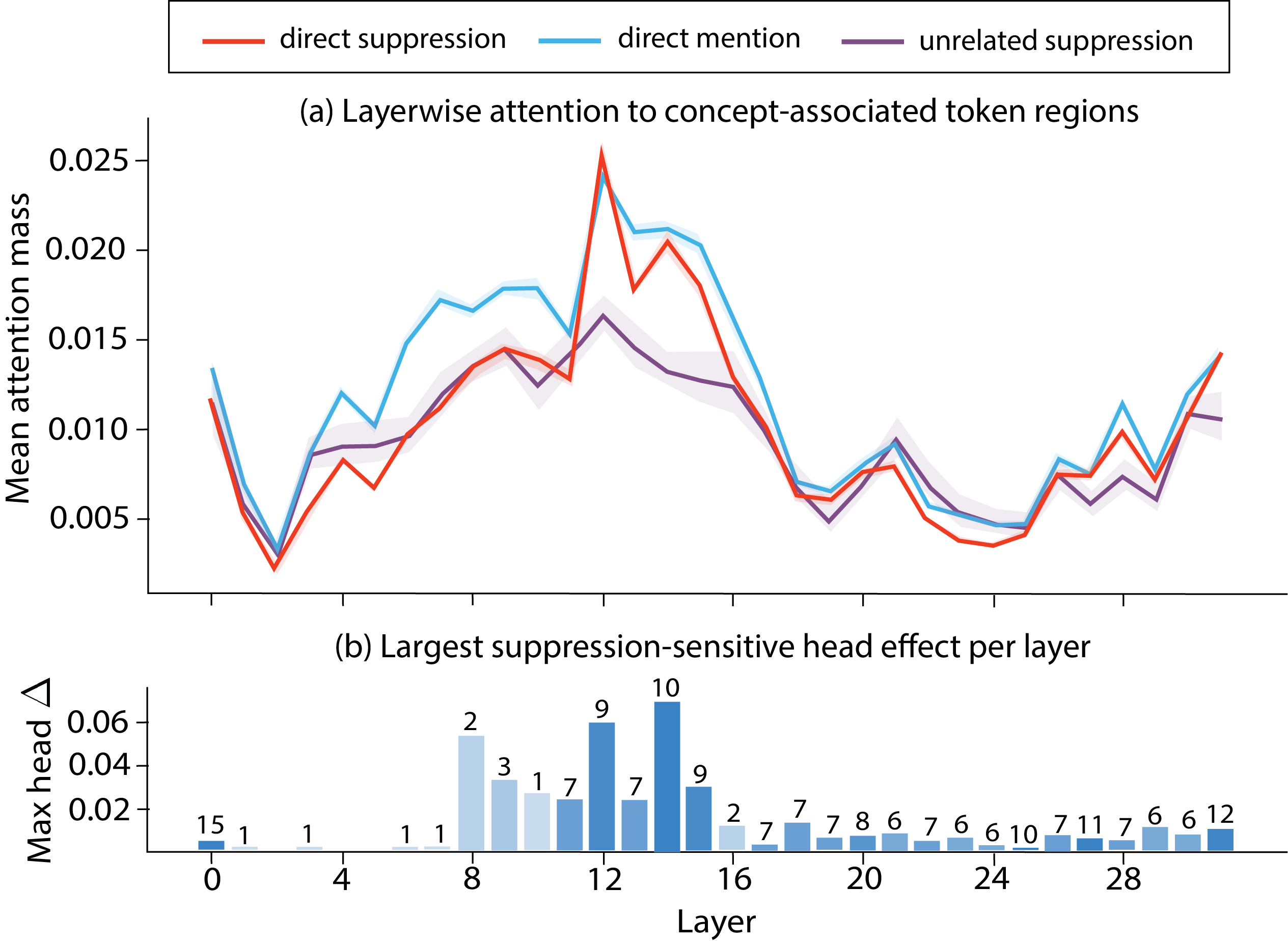}
    \caption{\textbf{Attention allocation.} Aggregate layerwise attention differences were modest overall (top), but head specialization emerged in middle layers (bottom). Bar heights indicate the largest positive head-level suppression effect within each layer. Numbers above bars denote the number of heads with significantly positive suppression effects (\(95\%\) CIs entirely above zero).}
    \label{fig:attention_dynamics}
    \vspace{-15pt}
\end{figure}

Although aggregate attention decreased, this average masked substantial head-level heterogeneity. A subset of middle-layer heads allocated greater attention to concept-associated regions under suppression than under direct mention (Fig.~\ref{fig:attention_dynamics}). The strongest effect emerged in layer 14 head 4, where suppression produced markedly elevated attention to target-alias regions ($\Delta = 0.068$, $95\%$ CI $[0.061, 0.076]$, $d = 1.73$, $p < .001$). These results suggest that suppression-related routing is localized rather than globally distributed.

\subsection{Semantic Control Analyses}
One may argue that suppression prompts globally alter transformer attention patterns independent of the prohibited concept itself, such that the observed routing effects simply reflect generic refusal behavior. To evaluate this possibility, we compared direct suppression against unrelated suppression controls in which the model was instructed to suppress a different concept while attention was still measured toward the original target-associated token regions.

At the aggregate level, direct suppression produced only modestly greater target-associated routing than unrelated suppression controls ($\Delta=9.45 \times 10^{-4}$, $t(6)=0.63$, $p=.55$), indicating that global attention allocation differences alone do not strongly distinguish concept-specific suppression from generic refusal dynamics.

However, headwise analyses revealed a different pattern. Several middle and late-layer attention heads exhibited large and highly reliable increases in routing toward prohibited concept regions under suppression relative to direct mention conditions, with several heads showing very large effect sizes ($d>1.0$; Appendix \ref{app:table-exp2-control}). These findings suggest that suppression-related processing is concentrated within specialized attention circuitry rather than uniformly distributed across the transformer stack.

\section{Experiment 3: Behavioral Leakage}\label{exp3}
The previous experiments demonstrate that suppressed concepts remain both representationally recoverable and computationally active. We next investigate whether these latent representations continue to influence downstream generation behavior. 

Rather than measuring explicit forbidden-token production alone, we evaluate whether suppression prompts increase semantic similarity between generated outputs and prohibited concepts despite successful lexical avoidance. This allows us to detect indirect semantic leakage through concept-adjacent associations. If suppression primarily constrains surface-level expression while preserving latent semantic activation, then generated outputs under suppression should remain measurably closer to prohibited concepts than matched absent baselines.




\subsection{Results}

\begin{figure}[t]
    \centering
    \includegraphics[width=0.95\linewidth]{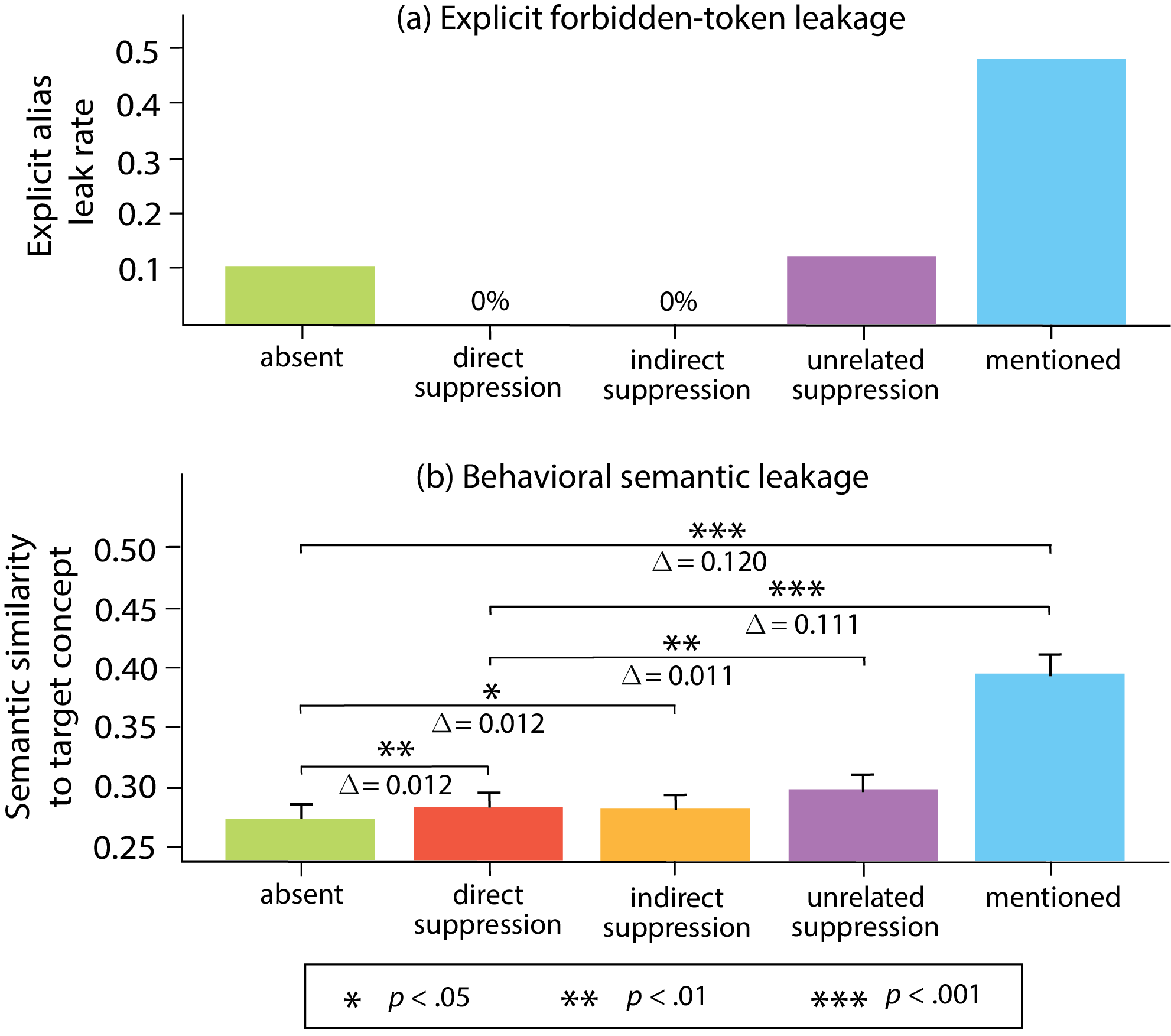}
    \caption{\textbf{Behavioral and semantic leakage.}
    While suppression largely eliminated explicit concept mentions (top), suppressed outputs remained significantly more associated with the prohibited concepts than concept-absent baselines (bottom). Examples of the prompt conditions are provided in Appendix~\ref{app:prompts}; complete specifications are available in the code repository. Error bars denote 95\% CIs.}
    \label{fig:exp3_behavioral}
    \vspace{-15pt}
\end{figure}

We first evaluated whether suppression instructions successfully prevented explicit lexical realization of prohibited concepts. As expected, direct mention prompts produced frequent forbidden-token leakage ($48.3\%$), whereas both direct and indirect suppression reduced explicit alias leakage to zero (Fig.~\ref{fig:exp3_behavioral}). In contrast, absent and unrelated suppression conditions exhibited only low background rates of spontaneous concept mention ($10.0\%$ and $12.3\%$, respectively). These results indicate that suppression instructions were behaviorally effective at the lexical level. 

Despite successful lexical suppression, prohibited concepts continued to influence downstream generations semantically. Direct suppression produced significantly greater semantic similarity to target concepts than concept-absent baselines ($\Delta = 0.012$, $95\%$ CI $[0.003, 0.021]$, $d = 0.23$, $p = .008$). Although substantially weaker than direct mention effects (Appendix \ref{app:table-exp3}), suppression outputs nevertheless remained measurably shifted toward prohibited semantic content.

Indirect semantic suppression produced a comparable effect despite never explicitly naming the prohibited concept. Relative to absent baselines, indirect suppression increased semantic similarity toward prohibited concepts ($\Delta = 0.012$, $95\%$ CI $[0.001, 0.022]$, $d = 0.19$, $p = .027$). This suggests that semantic leakage was not solely driven by direct lexical repetition within the prompt itself.



Semantic drift also varied across concepts. Concepts like ``knife,'' ``password,'' and ``self-harm'' exhibited stronger drift under suppression than other concepts (Appendix~\ref{app:table-exp3}). This heterogeneity suggests that semantic persistence may depend on the broader representational organization and contextual associations for individual concepts.



\section{Experiment 4: Cross-Model Analysis}\label{exp4}
The previous experiments demonstrated that suppressed concepts remain recoverable, computationally active, and behaviorally influential in Llama 3.1-8B. We next evaluate whether these effects generalize across transformer architectures by replicating the analysis in Mistral-7B and Gemma-7B-IT.

Cross-model analyses used indirect prompting with mean pooling over non-padding tokens. Because target concepts were never explicitly named within the prompts themselves, consistent effects across models would suggest that suppression-related latent persistence reflects a broader property of aligned transformer language models rather than an artifact of a particular model family.

\subsection{Results}

\begin{figure}[t]
    \centering
    \includegraphics[width=0.95\linewidth]{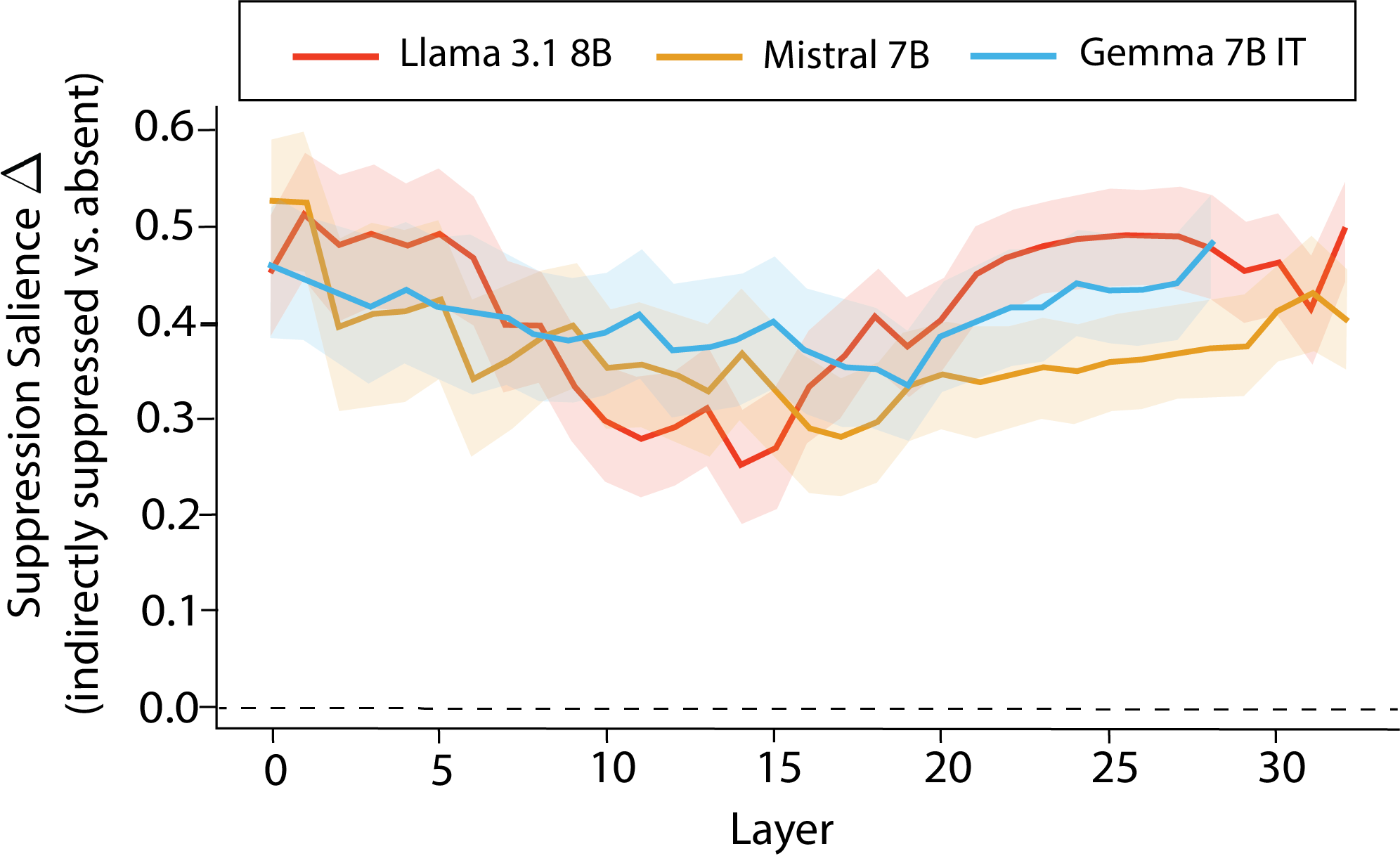}
    
    \caption{\textbf{Cross-model replication.}
Elevated recoverability under indirect suppression generalizes across architectures, suggesting that suppression-related semantic persistence is not model-specific.}
    \label{fig:exp4_layers}
    \vspace{-20pt}
\end{figure}

Probe performance remained high across models, with mean probe $AUC$s \(\geq 0.900\) and mean accuracies \(\geq 0.806\). Peak indirect suppression salience values were similarly large across model families (\(\Delta = 0.481\)--\(0.526\), all \(p < .001\), \(d = 1.25\)--\(1.53\)). Per-model statistics are provided in Appendix~\ref{app:stats-4}.

All three architectures exhibited a consistent non-monotonic structure in which suppression salience was significantly reduced in middle layers relative to both early and late layers (all \(p \leq .015\)). The effect was strongest in Llama (\(\Delta = 0.099\)--\(0.136\)), followed by Mistral (\(\Delta = 0.047\)--\(0.084\)) and Gemma (\(\Delta = 0.039\)).


Although the overall non-monotonic structure replicated consistently, the magnitude of late-layer persistence varied across architectures. Llama exhibited significantly stronger late-layer suppression salience than both Mistral (\(\Delta = 0.097\), \(p < .001\)) and Gemma (\(\Delta = 0.054\), \(p = .0014\)). An ordinary least squares interaction analysis further supported the presence of model-by-region effects, \(F(8,86)=11.78\), \(p < .001\). 

Nevertheless, all evaluated models exhibited the same statistically reliable ordering observed throughout the preceding experiments. That is, direct mention produced the strongest latent recoverability, indirect suppression produced intermediate recoverability, and concept-absent prompts produced the weakest recoverability overall, with all pairwise contrasts significant under paired sign-flip permutation tests (\(p < .001\); Appendix~\ref{app:stats-4}).

\section{Discussion}
Across multiple experiments, our results suggest prohibited concepts frequently remain representationally recoverable even when models successfully avoid producing the forbidden concepts behaviorally. Interestingly, this persistence survived indirect semantic suppression conditions in which target concepts were never explicitly named, indicating that the observed recoverability cannot be explained solely by lexical overlap between the prompt and the prohibited concept.

One possible interpretation is that suppression prompts require models to internally maintain representations associated with the prohibited concept in order to monitor and avoid generating it. Under this view, successful constraint following may itself depend upon preserving latent concept information throughout decoding rather than eliminating it entirely. More generally, exclusionary instructions may increase the salience of concept-associated semantic manifolds because the model must continually evaluate candidate generations relative to the prohibited concept during autoregressive decoding.

Behavioral analyses further indicated that suppression influenced generation toward semantically proximate outputs despite successful exclusion of the forbidden terms. For example, when instructed not to mention self-harm, models frequently redirected generation toward themes of crisis management, safety planning, and psychiatric care while avoiding the prohibited term itself. Such outputs retained substantial semantic overlap with the suppressed concept, suggesting that persistent representations continue to influence downstream generation and are not just detectable artifacts of probing.



Attention analyses provide a tentative mechanistic account of this persistence. We do not interpret individual attention heads as encoding discrete concepts. Rather, suppression was associated with selective changes in attention allocation, particularly under indirect suppression, suggesting structured differences in information routing rather than a uniform increase in refusal-related computation.

\subsection{Implications for Alignment Evaluation}
These findings raise a broader methodological point for the evaluation of aligned language models. Benchmarking refusal behavior exclusively at the output level may obscure substantial internal semantic processing associated with prohibited concepts. Analyses of latent representations, attentional routing, and semantic persistence may therefore provide complementary insight into how aligned systems internally implement constraint-following behavior during generation.

More broadly, our work puts forth that behavioral alignment and internal representational suppression may constitute partially separable objectives. A model may successfully avoid producing prohibited outputs while nevertheless retaining substantially relevant latent structure associated with the suppressed concept. Understanding how these two forms of alignment interact becomes increasingly important as language models are deployed in long-horizon, open-ended, safety-critical settings.

\subsection{Taming the White Bear?}
We did not pursue a direct computational analogue of the classic human ``white bear'' effect in its strongest psychological form. Under ironic process theory, suppression prompting can paradoxically increase the accessibility of a target concept beyond ordinary activation levels. A direct analogue in language models would therefore predict:
\[
f_c(h_{\mathrm{suppressed}})
>
f_c(h_{\mathrm{mentioned}})
\]
where suppression produces greater recoverability than mentioning the concept without suppression.

Our experiments did not consistently support this stronger pattern at the global representational level. Instead, the dominant result across models and pooling schemes was a graded structure:
\[
f_c(h_{\mathrm{mentioned}})
>
f_c(h_{\mathrm{suppressed}})
\gg
f_c(h_{\mathrm{absent}})
\]
in which direct mention produced the strongest recoverability, while suppression preserved substantial but attenuated recoverability relative to explicit mention. Thus, the present findings are more appropriately interpreted as evidence of persistent semantic recoverability under suppression than a full computational realization of ironic hyperactivation. Nevertheless, localized attention heads and intermediate-layer routing patterns occasionally exhibited stronger suppression-related effects than direct mention, suggesting that stronger ``white bear''-like dynamics may emerge under more cognitively demanding or longer-horizon suppression settings.

\subsection{Limitations and Future Work}

We acknowledge several limitations of the present findings. First, linear probes show that information about a prohibited concept remains statistically recoverable from model activations; they do not imply that models store concepts in a discrete or symbolic form. Second, attention allocation alone should not be interpreted as a complete explanation of model reasoning or causal influence. Although our control conditions reduce several straightforward lexical and refusal-related confounds, semantic suppression prompts may still introduce subtle distributional biases that contribute to the observed effects. Third, the present experiments focus primarily on instruction-tuned autoregressive language models and may not generalize uniformly across other architectures or non-autoregressive generation paradigms.

Finally, while the current analyses establish robust correlational evidence for persistent semantic recoverability under suppression, they do not determine the causal role these representations play during generation itself. An important direction for future work is to determine whether persistent concept representations actively contribute to successful constraint following or merely accompany it. Approaches based on causal interventions, activation steering, and circuit-level analysis may help distinguish between these possibilities. It also remains unclear whether stronger white-bear-like effects emerge in more complex reasoning tasks, or settings requiring sustained suppression over extended contexts or long-horizon interactions. More broadly, understanding how suppression is implemented within distributed representations may provide a path toward more mechanistically grounded approaches to alignment and controllability.

\appendix

\section*{Appendices}
Appendix~\ref{app:concepts} describes the concept library used in our experiments. Appendix~\ref{app:prompts} presents representative prompt templates and conditions. Appendix~\ref{app:stats} reports additional descriptive statistics corresponding to the analyses discussed in the main text.

\section{Concept Library} \label{app:concepts}
Our concept library is implemented as a lightweight JSON-based framework in which each concept is represented through multiple complementary views. These views include direct aliases, indirect semantic descriptions, contextual elicitation prompts, positive examples, matched negatives, and semantically adjacent hard negatives. The final library contains 17 concept categories spanning safety-relevant, socially sensitive, and identity-related domains (e.g., weapons, violence, self-harm, politics, religion, privacy, fraud, and copyrighted material). In all, the library comprises 113 aliases, 102 indirect descriptions, 136 contextual prompts, 408 positive examples, 408 matched negatives, and 170 hard negatives (986 entries total).



\begin{lstlisting}[
language=json,
basicstyle=\ttfamily\footnotesize,
breaklines=true,
breakatwhitespace=true,
columns=fullflexible,
keepspaces=true,
showstringspaces=false,
frame=single,
numbers=none,
caption={Abbreviated concept specification example.}
]
"white_bear": {
  "aliases": [
    "white bear", "polar bear", "arctic bear"
  ],

  "indirect_descriptions": [
    "large white mammal in arctic regions",
    "predator associated with sea ice"
  ],

  "contexts": [
    "Describe an arctic environment.",
    "Describe animals living near sea ice."
  ],

  "positive": [
    "The white bear walked across the frozen landscape.",
    "A polar bear stood near the ice shelf."
  ],

  "negative": [
    "The seal rested beside the frozen shoreline.",
    "Snow drifted across the empty ice field."
  ],

  "negative_hard": [
    "A large animal moved through the snow.",
    "The arctic predator searched near the shoreline."
  ]
}, ...
\end{lstlisting}

\section{Prompt Conditions and Templates} \label{app:prompts}
Each concept–context pair was evaluated under five matched prompting conditions designed to isolate suppression effects while preserving contextual structure. Across 17 concepts and 136 contextual prompts, the framework instantiated 680 prompt instances spanning absent, mentioned, direct suppression, indirect suppression, and unrelated suppression control conditions. Table~\ref{tab:prompt_conditions} summarizes each condition and provides representative prompts.

\begin{table*}[t]
\centering
\small
\caption{\textbf{Prompt conditions used throughout the experiments.} Each concept--context pair was instantiated under five matched conditions that varied only in the suppression instruction while preserving the underlying context.}
\label{tab:prompt_conditions}
\begin{tabular}{l l p{0.30\linewidth} p{0.38\linewidth}}
\toprule
\textbf{} & \textbf{Condition} & \textbf{Description} & \textbf{Example Prompt Instantiations} \\
& & & Prefix: \texttt{Describe an arctic environment.} \\
\midrule

$\mathrm{abs}$
& Absent
& Baseline context without the target concept
& --- \\

$\mathrm{men}$
& Direct Mention
& Explicit inclusion of the target concept
& \texttt{Include white bear in the description.} \\

$\mathrm{sup}$
& Direct Suppression
& Direct instruction prohibiting the target concept
& \texttt{Do not mention white bear.} \\

$\mathrm{ind}$
& Indirect Suppression
& Prohibition of semantically descriptive paraphrases
& \texttt{Do not mention or allude to a large white mammal living in arctic regions.} \\

$\mathrm{ctrl}$
& Unrelated Suppression
& Suppression instruction targeting an unrelated concept
& \texttt{Do not mention flowers.} \\

\bottomrule
\end{tabular}
\end{table*}


\section{Raw Means and Statistical Tables}\label{app:stats}
This appendix provides supporting quantitative results for all experiments. We report probe quality metrics, raw condition means, layerwise recoverability statistics, attention allocation analyses, behavioral leakage measurements, and cross-model replication results. These tables are intended to complement the summarized findings reported in the main text by providing the underlying descriptive statistics, effect sizes, confidence intervals, and significance tests used throughout the paper.

\subsection{Experiment 1: Recoverability}\label{app:stats-1}
Table~\ref{tab:probe_quality} reports probe quality across pooling strategies, Table~\ref{tab:raw_salience_means} reports the peak-layer raw condition means for the primary suppression comparison, and Table~\ref{tab:exp1_raw_means} reports full layerwise probe recoverability statistics for the pooling configuration used.

\begin{table}[H]
\centering
\small
\caption{\textbf{Probe quality under pooling strategies}. Mean probe quality across layers, concepts, and split seeds.}
\label{tab:probe_quality}
\begin{tabular}{lccc}
\toprule
\textbf{Pooling} & Accuracy ($acc.$) & \textbf{$AUC$} & \textbf{$F_1$} \\
\midrule
Target tokens & 0.828 & 0.880 & 0.796 \\
Mean nonpad & 0.827 & 0.919 & 0.820 \\
Last nonpad & 0.779 & 0.860 & 0.754 \\
Indirect only & 0.827 & 0.919 & 0.820 \\
\bottomrule
\end{tabular}
\vspace{-10pt}
\end{table}


\begin{table*}[t]
\centering
\small
\caption{
\textbf{Peak suppression effects across activation pooling strategies.}
Reported values correspond to the layer exhibiting the largest paired suppression-related recoverability difference for each pooling method. For direct suppression rows, $\Delta = \mu_{\mathrm{sup}} - \mu_{\mathrm{abs}}$. For indirect suppression, $\Delta = \mu_{\mathrm{ind}} - \mu_{\mathrm{abs}}$.
}
\label{tab:raw_salience_means}

\begin{tabular}{lccccccc}
\toprule
\textbf{Pooling}
& Layer
& $\mu_{\mathrm{abs}}$
& $\mu_{\mathrm{sup}}$
& $\mu_{\mathrm{ind}}$
& $\mu_{\mathrm{men}}$
& $\Delta$
& Cohen's $d$ \\
\midrule

Target tokens
& 4
& 0.073
& 0.883
& 0.556
& 0.873
& 0.810
& 2.663 \\

Mean nonpad
& 1
& 0.160
& 0.792
& 0.664
& 0.761
& 0.632
& 2.112 \\

Last nonpad
& 3
& 0.389
& 0.667
& 0.604
& 0.366
& 0.278
& 0.722 \\

\midrule

Indirect suppression
& 32
& 0.138
& 0.846
& 0.668
& 0.870
& 0.529
& 1.745 \\

\bottomrule
\end{tabular}
\end{table*}

\begin{table*}[t]
\centering
\small
\caption{
\textbf{Raw probe recoverability scores by layer under target-token pooling.} Reported values correspond to paired means across all concepts and contexts.
}
\label{tab:exp1_raw_means}
\begin{tabular}{ccccccc}
\toprule
\textbf{Layer}
& $\mu_{\mathrm{abs}}$
& $\mu_{\mathrm{sup}}$
& $\mu_{\mathrm{men}}$
& $\Delta_{\mathrm{sup-abs}}$
& $\Delta_{\mathrm{sup-men}}$
& Cohen's $d$ \\
\midrule

0  & 0.219 & 0.802 & 0.802 & 0.583 & 0.000 & 1.956 \\
4  & 0.073 & 0.883 & 0.873 & 0.810 & 0.010 & 2.663 \\
8  & 0.157 & 0.816 & 0.883 & 0.659 & -0.067 & 1.701 \\
12 & 0.140 & 0.830 & 0.903 & 0.689 & -0.074 & 2.163 \\
16 & 0.260 & 0.829 & 0.895 & 0.569 & -0.066 & 1.508 \\
20 & 0.237 & 0.863 & 0.901 & 0.625 & -0.038 & 1.584 \\
24 & 0.224 & 0.854 & 0.881 & 0.630 & -0.027 & 1.619 \\
28 & 0.279 & 0.862 & 0.883 & 0.583 & -0.021 & 1.526 \\
32 & 0.258 & 0.846 & 0.870 & 0.588 & -0.024 & 1.518 \\

\bottomrule
\end{tabular}
\end{table*}

\subsection{Experiment 2: Attention Dynamics} \label{app:table-exp2-control}

Table~\ref{tab:exp2_global_comparisons} summarizes attention allocation across conditions. Table~\ref{tab:exp2_top_heads} reports the largest head-level suppression effects relative to direct mention. Table~\ref{tab:table-exp2-control} reports the strongest heads differentiating direct suppression from unrelated suppression controls.

\begin{table*}[t]
\centering
\small
\caption{\textbf{Global attention allocation comparisons across conditions.} 
Attention values correspond to mean attention mass allocated toward target-associated token regions. Confidence interveals are over paired differences.}
\label{tab:exp2_global_comparisons}
\begin{tabular}{lccccccc}
\toprule
Comparison & $\mu_A$ & $\mu_B$ & $\Delta$ & 95\% CI & Cohen's $d$ & Wilcoxon $p$ & $n$ \\
\midrule
$\mathrm{sup}$ $-$ $\mathrm{men} $
& 0.00995 
& 0.01189 
& $-0.00194$ 
& [$-0.00213$, $-0.00175$] 
& $-1.73$ 
& $1.53\times10^{-21}$ 
& 129 \\

$\mathrm{sup}$ $-$ $\mathrm{ctrl}$
& 0.01040 
& 0.00945 
& $9.45\times10^{-4}$ 
& [$-1.61\times10^{-3}$, $3.58\times10^{-3}$] 
& 0.24 
& 0.8125 
& 7 \\
\bottomrule
\end{tabular}
\end{table*}

\begin{table*}[t]
\centering
\small
\caption{\textbf{Largest suppression-sensitive attention heads.}
Headwise effects computed as attention allocation toward target-associated token regions under direct suppression minus direct mention conditions.}
\label{tab:exp2_top_heads}
\begin{tabular}{cccccc}
\toprule
Layer & Head & $\Delta$ & 95\% CI & Cohen's $d$ & Wilcoxon $p$ \\
\midrule
14 & 4  & 0.0684 & [0.0614, 0.0756] & 1.73 & $6.51\times10^{-23}$ \\
12 & 29 & 0.0588 & [0.0530, 0.0647] & 1.76 & $6.51\times10^{-23}$ \\
12 & 6  & 0.0526 & [0.0452, 0.0598] & 1.23 & $7.49\times10^{-23}$ \\
8  & 14 & 0.0525 & [0.0450, 0.0603] & 1.19 & $8.42\times10^{-23}$ \\
9  & 13 & 0.0322 & [0.0282, 0.0363] & 1.35 & $1.01\times10^{-22}$ \\
15 & 13 & 0.0289 & [0.0249, 0.0331] & 1.24 & $1.06\times10^{-22}$ \\
10 & 28 & 0.0264 & [0.0232, 0.0298] & 1.35 & $6.51\times10^{-23}$ \\
11 & 8  & 0.0234 & [0.0200, 0.0270] & 1.12 & $1.19\times10^{-22}$ \\
13 & 19 & 0.0233 & [0.0204, 0.0262] & 1.35 & $6.83\times10^{-23}$ \\
12 & 3  & 0.0190 & [0.0168, 0.0213] & 1.46 & $9.03\times10^{-23}$ \\
\bottomrule
\end{tabular}
\end{table*}

\begin{table*}[t]
\centering
\small
\caption{\textbf{Top attention heads differentiating direct suppression from unrelated suppression controls.}
Attention differences are computed as mean attention allocation toward target-associated token regions under direct suppression minus unrelated suppression.}
\label{tab:table-exp2-control}
\begin{tabular}{cccccc}
\toprule
Layer & Head & $\Delta$ & 95\% CI & Cohen's $d$ & $p$ \\
\midrule
14 & 4  & 0.0531 & [0.0306, 0.0741] & 1.59 & 0.0057 \\
12 & 29 & 0.0407 & [0.0214, 0.0623] & 1.36 & 0.0115 \\
20 & 3  & 0.0377 & [0.0229, 0.0537] & 1.67 & 0.0044 \\
14 & 19 & 0.0336 & [0.0210, 0.0491] & 1.59 & 0.0057 \\
9  & 13 & 0.0321 & [0.0155, 0.0503] & 1.26 & 0.0159 \\
\bottomrule
\end{tabular}
\end{table*}

\subsection{Experiment 3: Behavioral Leakage} \label{app:table-exp3}
\begin{figure}[H]
    \centering
    \includegraphics[width=\linewidth]{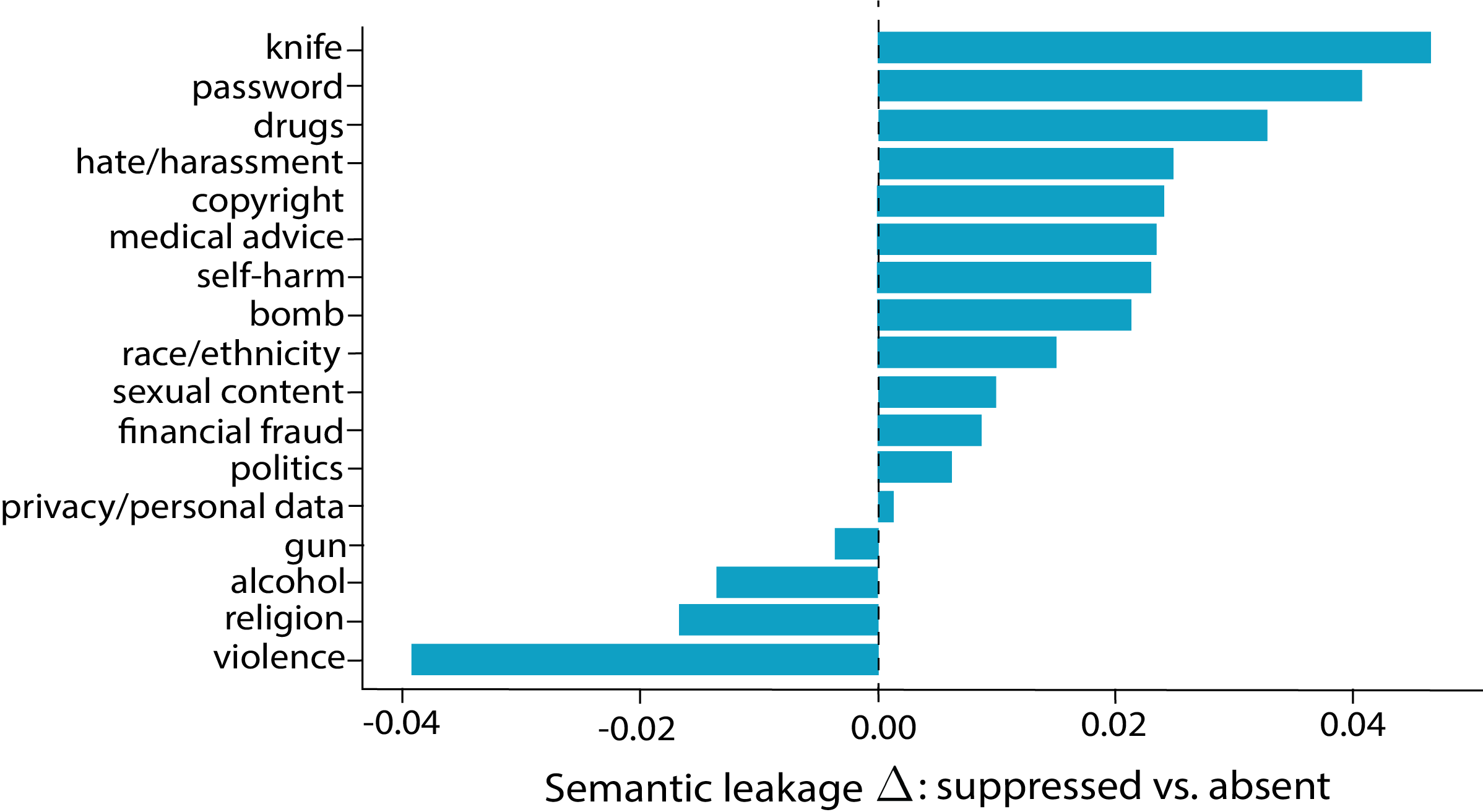}
    \caption{\textbf{Suppression-related semantic drift by concept category.} Values indicate the mean difference in semantic similarity between suppression and concept-absent conditions. Positive values denote greater residual association with the prohibited concept.}
    \label{fig:semantic_drift}
\end{figure}

Table~\ref{tab:behavioral_leakage} summarizes semantic similarity and leakage rates across conditions. Table~\ref{tab:behavioral_pairwise} reports pairwise comparisons between conditions. Fig.~\ref{fig:semantic_drift} reports semantic drift by concept category. 

\begin{table*}[t]
\centering
\small
\caption{\textbf{Behavioral leakage by condition.} Values summarize semantic similarity between outputs and prohibited concepts, together with explicit alias leakage rates. Confidence intervals are over mean semantic similarity.}
\label{tab:behavioral_leakage}
\begin{tabular}{lccccc}
\toprule
Condition & Mean similarity & SEM & 95\% CI & Explicit leak rate & $n$ \\
\midrule
$\mathrm{abs}$ & 0.275 & 0.006 & [0.264, 0.286] & 0.100 & 408 \\
$\mathrm{men}$ & 0.395 & 0.007 & [0.380, 0.410] & 0.483 & 408 \\
$\mathrm{sup}$ & 0.284 & 0.006 & [0.273, 0.296] & 0.000 & 375 \\
$\mathrm{ind}$ & 0.282 & 0.006 & [0.270, 0.294] & 0.000 & 380 \\
$\mathrm{ctrl}$ & 0.299 & 0.006 & [0.286, 0.311] & 0.123 & 408 \\
\bottomrule
\end{tabular}
\end{table*}

\begin{table*}[t]
\centering
\small
\caption{\textbf{Pairwise behavioral leakage comparisons.} Paired differences are computed over matched concept--context items. Positive $\Delta$ values indicate greater semantic similarity in condition A than condition B.}
\label{tab:behavioral_pairwise}
\begin{tabular}{lcccccc}
\toprule
Comparison & $\mu_A$ & $\mu_B$ & $\Delta$ & 95\% CI & Cohen's $d$ & $p_{\mathrm{Wilcoxon}}$ \\
\midrule
$\mathrm{sup}$ $-$ $\mathrm{abs}$ & 0.285 & 0.273 & 0.0120 & [0.0035, 0.0211] & 0.234 & 0.005 \\
$\mathrm{men}$ $-$ $\mathrm{abs}$ & 0.395 & 0.275 & 0.1204 & [0.1012, 0.1411] & 1.015 & $6.83\times 10^{-22}$ \\
$\mathrm{sup}$ $-$ $\mathrm{men}$ & 0.285 & 0.396 & -0.1105 & [-0.1299, -0.0920] & -1.004 & $5.13\times 10^{-21}$ \\
$\mathrm{sup}$ $-$ $\mathrm{ctrl}$ & 0.285 & 0.297 & -0.0115 & [-0.0191, -0.0041] & -0.258 & 0.001 \\
$\mathrm{ind}$ $-$ $\mathrm{abs}$ & 0.285 & 0.273 & 0.0115 & [0.0011, 0.0217] & 0.194 & 0.008 \\
\bottomrule
\end{tabular}
\end{table*}

\subsection{Experiment 4: Cross-Model Analysis}\label{app:stats-4}

\begin{figure}[H]
    \centering
    \includegraphics[width=\linewidth]{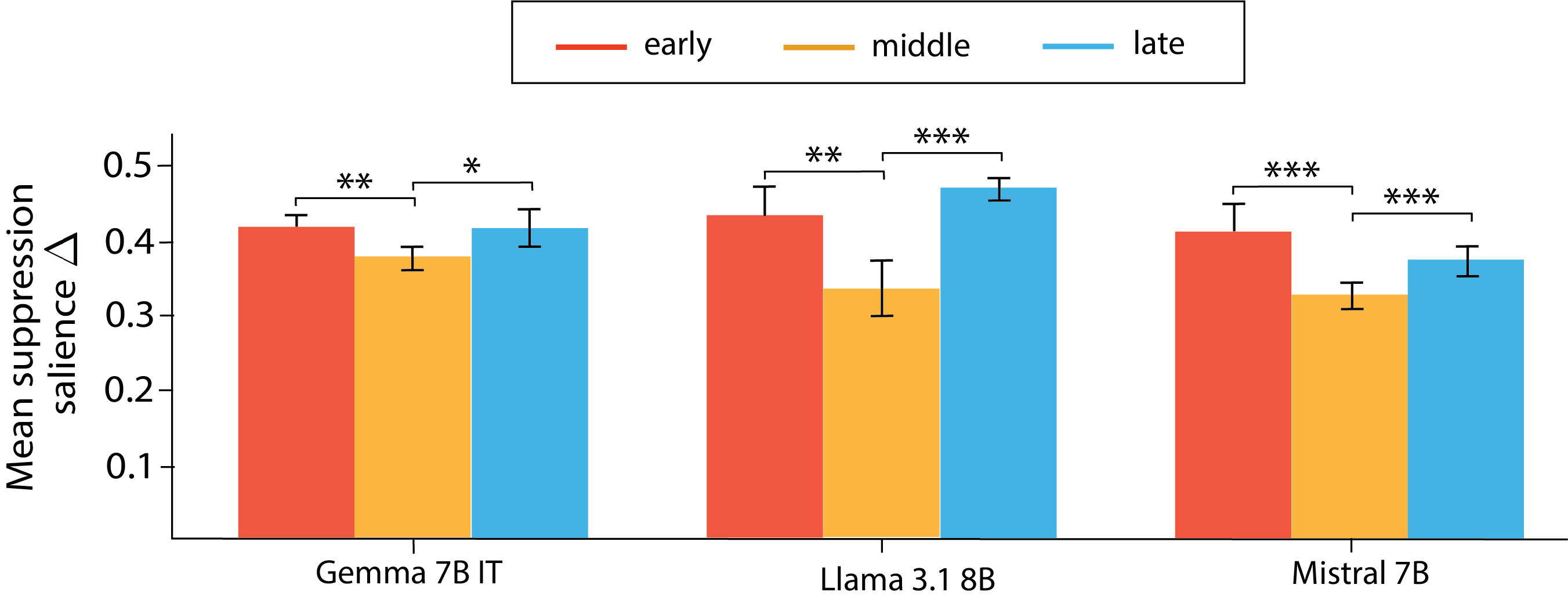}
    \caption{
\textbf{Layer-region suppression salience across model families.} All models exhibited significantly lower suppression salience in middle layers than in early or late layers, revealing a non-monotonic suppression profile across transformer architectures.
}
    \label{fig:cross_model_regions}
    \vspace{-15pt}
\end{figure}

\begin{figure}[H]
    \centering
    \includegraphics[width=\linewidth]{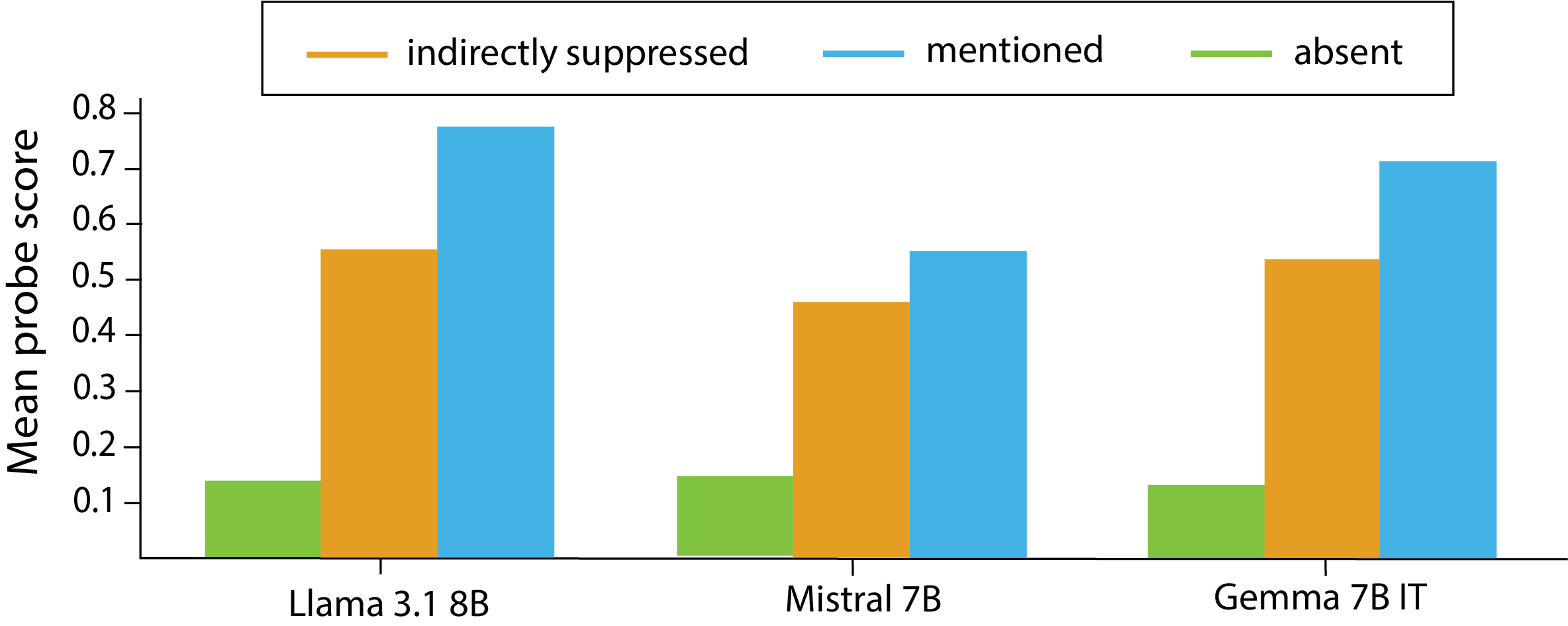}
    \caption{
\textbf{Condition ordering across model families.}
Mean probe scores under concept-absent, indirect suppression, and direct mention conditions across models. All pairwise contrasts were significant at \(p < .001\).
}
    \label{fig:exp4_ordering}
\end{figure}

Table~\ref{tab:cross-model-suppression} summarizes probe performance and peak suppression-salience statistics across model families. Table~\ref{tab:layer-region-summary} reports layer-region suppression salience, illustrated by Fig.~\ref{fig:cross_model_regions}. Table~\ref{tab:condition-ordering} reports condition-level recoverability across architectures, depicted as Fig.~\ref{fig:exp4_ordering}. 

\begin{table*}[t]
\centering
\small
\caption{
\textbf{Cross-model replication of indirect suppression salience.}
Probe performance and peak suppression effects across model families under indirect-only suppression prompts. Peak \(\Delta\) denotes the maximum layerwise suppression salience difference relative to concept-absent baselines. All reported effects were significant (\(p < .001\)).
}
\label{tab:cross-model-suppression}
\begin{tabular}{lcccccccc}
\toprule
Model & Layers & $AUC$ & $acc.$ & $F_1$ & Peak layer & Peak $\Delta$ & 95\% CI & $d$ \\
\midrule
\texttt{Llama-3-8B} & 33 & 0.916 & 0.824 & 0.816 & 1  & 0.511 & [0.444, 0.577] & 1.25 \\
\texttt{Mistral-7B} & 33 & 0.900 & 0.806 & 0.798 & 0  & 0.526 & [0.455, 0.591] & 1.33 \\
\texttt{Gemma-7B-IT}     & 29 & 0.910 & 0.811 & 0.806 & 28 & 0.481 & [0.426, 0.535] & 1.53 \\
\bottomrule
\end{tabular}

\end{table*}

\begin{table*}[t]
\centering
\small
\caption{
\textbf{Layer-region summary of indirect suppression salience across model families.} Mean $\Delta$ denotes the average layerwise suppression salience difference relative to concept-absent baselines under indirect-only suppression prompts.
}
\label{tab:layer-region-summary}
\begin{tabular}{llcccc}
\toprule
Model & Region & Layers & Mean $\Delta$ & 95\% CI & Range \\
\midrule
\texttt{Gemma-7B-IT}     & Early  & 10 & 0.418 & [0.404, 0.433] & [0.381, 0.459] \\
\texttt{Gemma-7B-IT}     & Middle & 9  & 0.379 & [0.368, 0.391] & [0.353, 0.409] \\
\texttt{Gemma-7B-IT}     & Late   & 10 & 0.418 & [0.393, 0.440] & [0.336, 0.481] \\
\midrule
\texttt{Llama-3-8B} & Early  & 11 & 0.435 & [0.393, 0.472] & [0.295, 0.511] \\
\texttt{Llama-3-8B} & Middle & 11 & 0.336 & [0.301, 0.372] & [0.250, 0.446] \\
\texttt{Llama-3-8B} & Late   & 11 & 0.472 & [0.457, 0.484] & [0.412, 0.496] \\
\midrule
\texttt{Mistral-7B} & Early  & 11 & 0.412 & [0.379, 0.449] & [0.343, 0.526] \\
\texttt{Mistral-7B} & Middle & 11 & 0.328 & [0.311, 0.343] & [0.280, 0.367] \\
\texttt{Mistral-7B} & Late   & 11 & 0.375 & [0.361, 0.392] & [0.344, 0.430] \\
\bottomrule
\end{tabular}
\end{table*}

\begin{table*}[t]
\centering
\small
\caption{
\textbf{Condition-ordering analysis across model families under indirect-only suppression prompts.} Pairwise differences were evaluated using paired sign-flip permutation tests and bootstrap confidence intervals over matched concept, context, and layer representations. All pairwise comparisons were significant at \(p < .001\).
}
\label{tab:condition-ordering}
\begin{tabular}{lccccccccc}
\toprule
Model 
& $\mathrm{abs}$ 
& $\mathrm{ind}$
& $\mathrm{men}$
& $\mathrm{ind}-\mathrm{abs}$
& 95\% CI 
& $d$ 
& $\mathrm{men}$-$\mathrm{ind}$
& 95\% CI 
& $d$ \\
\midrule

\texttt{Llama-3-8B}
& 0.138
& 0.553
& 0.772
& 0.414
& [0.404, 0.425]
& 1.15
& 0.219
& [0.208, 0.231]
& 0.57 \\

\texttt{Mistral-7B}
& 0.088
& 0.459
& 0.550
& 0.372
& [0.360, 0.383]
& 0.94
& 0.091
& [0.081, 0.101]
& 0.28 \\

\texttt{Gemma-7B-IT}
& 0.131
& 0.537
& 0.708
& 0.406
& [0.394, 0.418]
& 1.03
& 0.171
& [0.160, 0.181]
& 0.50 \\

\bottomrule
\end{tabular}

\end{table*}

\begin{thebibliography}{26}
\providecommand{\natexlab}[1]{#1}

\bibitem[{Bai et~al.(2022)Bai, Kadavath, Kundu, Askell, Kernion, Jones, Chen, Goldie, Mirhoseini, McKinnon et~al.}]{bai2022constitutional}
Yuntao Bai, Saurav Kadavath, Sandipan Kundu, Amanda Askell, Jackson Kernion, Andy Jones, Anna Chen, Anna Goldie, Azalia Mirhoseini, Cameron McKinnon, and 1 others. 2022.
\newblock Constitutional ai: Harmlessness from ai feedback.
\newblock \emph{arXiv preprint arXiv:2212.08073}.

\bibitem[{Belinkov(2022)}]{belinkov2022probing}
Yonatan Belinkov. 2022.
\newblock Probing classifiers: Promises, shortcomings, and advances.
\newblock \emph{Computational Linguistics}, 48(1):207--219.

\bibitem[{Burns et~al.(2022)Burns, Ye, Klein, and Steinhardt}]{burns2022discovering}
Collin Burns, Haotian Ye, Dan Klein, and Jacob Steinhardt. 2022.
\newblock Discovering latent knowledge in language models without supervision.
\newblock \emph{arXiv preprint arXiv:2212.03827}.

\bibitem[{Dennett(1984)}]{dennett1984cognitive}
Daniel~C Dennett. 1984.
\newblock \emph{Cognitive wheels: the frame problem of AI}.
\newblock Minds, Machines and Evolution. Cambridge University Press Cambridge, UK:.

\bibitem[{Elhage et~al.(2021)Elhage, Nanda, Olsson, Henighan, Joseph, Mann, Askell, Bai, Chen, Conerly, DasSarma, Drain, Ganguli, Hatfield-Dodds, Hernandez, Jones, Kernion, Lovitt, Ndousse, Amodei, Brown, Clark, Kaplan, McCandlish, and Olah}]{elhage2021mathematical}
Nelson Elhage, Neel Nanda, Catherine Olsson, Tom Henighan, Nicholas Joseph, Ben Mann, Amanda Askell, Yuntao Bai, Anna Chen, Tom Conerly, Nova DasSarma, Dawn Drain, Deep Ganguli, Zac Hatfield-Dodds, Danny Hernandez, Andy Jones, Jackson Kernion, Liane Lovitt, Kamal Ndousse, and 6 others. 2021.
\newblock A mathematical framework for transformer circuits.
\newblock \emph{Transformer Circuits Thread}.
\newblock Https://transformer-circuits.pub/2021/framework/index.html.

\bibitem[{Ettinger(2020)}]{ettinger2020bert}
Allyson Ettinger. 2020.
\newblock What {BERT} is not: Lessons from a new suite of psycholinguistic diagnostics for language models.
\newblock \emph{Transactions of the Association for Computational Linguistics}, 8:34--48.

\bibitem[{Geva et~al.(2023)Geva, Bastings, Filippova, and Globerson}]{geva2023dissecting}
Mor Geva, Jasmijn Bastings, Katja Filippova, and Amir Globerson. 2023.
\newblock Dissecting recall of factual associations in auto-regressive language models.
\newblock In \emph{Proceedings of the 2023 Conference on Empirical Methods in Natural Language Processing}, pages 12216--12235.

\bibitem[{Grattafiori et~al.(2024)Grattafiori, Dubey, Jauhri, Pandey, Kadian, Al-Dahle, Letman, Mathur, Schelten, Vaughan et~al.}]{grattafiori2024llama}
Aaron Grattafiori, Abhimanyu Dubey, Abhinav Jauhri, Abhinav Pandey, Abhishek Kadian, Ahmad Al-Dahle, Aiesha Letman, Akhil Mathur, Alan Schelten, Alex Vaughan, and 1 others. 2024.
\newblock The llama 3 herd of models.
\newblock \emph{arXiv preprint arXiv:2407.21783}.

\bibitem[{Jiang et~al.(2023)Jiang, Sablayrolles, Mensch, Bamford, Chaplot, de~las Casas, Bressand, Lengyel, Lample, Saulnier, Lavaud, Lachaux, Stock, Scao, Lavril, Wang, Lacroix, and Sayed}]{jiang2023mistral7b}
Albert~Q. Jiang, Alexandre Sablayrolles, Arthur Mensch, Chris Bamford, Devendra~Singh Chaplot, Diego de~las Casas, Florian Bressand, Gianna Lengyel, Guillaume Lample, Lucile Saulnier, Lélio~Renard Lavaud, Marie-Anne Lachaux, Pierre Stock, Teven~Le Scao, Thibaut Lavril, Thomas Wang, Timothée Lacroix, and William~El Sayed. 2023.
\newblock \href {https://arxiv.org/abs/2310.06825} {Mistral 7b}.
\newblock \emph{Preprint}, arXiv:2310.06825.

\bibitem[{Kassner and Sch{\"u}tze(2020)}]{kassner2020negated}
Nora Kassner and Hinrich Sch{\"u}tze. 2020.
\newblock Negated and misprimed probes for pretrained language models: Birds can talk, but cannot fly.
\newblock In \emph{Proceedings of the 58th annual meeting of the association for computational linguistics}, pages 7811--7818.

\bibitem[{Kumar et~al.(2026)Kumar, Gupta, Birur, Baswa, Agarwal, and Harshangi}]{kumar2026redirected}
Divyanshu Kumar, Ishita Gupta, Nitin~Aravind Birur, Tanay Baswa, Sahil Agarwal, and Prashanth Harshangi. 2026.
\newblock Redirected, not removed: Task-dependent stereotyping reveals the limits of {LLM} alignments.
\newblock \emph{arXiv preprint arXiv:2604.02669}.

\bibitem[{Li et~al.(2025)Li, Fu, Fan, Liu, Shu, Qin, Yang, King, and Ying}]{li2025implicit}
Jindong Li, Yali Fu, Li~Fan, Jiahong Liu, Yao Shu, Chengwei Qin, Menglin Yang, Irwin King, and Rex Ying. 2025.
\newblock Implicit reasoning in large language models: A comprehensive survey.
\newblock \emph{arXiv preprint arXiv:2509.02350}.

\bibitem[{McCoy et~al.(2019)McCoy, Pavlick, and Linzen}]{mccoy2019right}
R~Thomas McCoy, Ellie Pavlick, and Tal Linzen. 2019.
\newblock Right for the wrong reasons: Diagnosing syntactic heuristics in natural language inference.
\newblock In \emph{Proceedings of the 57th annual meeting of the association for computational linguistics}, pages 3428--3448.

\bibitem[{Meng et~al.(2022)Meng, Bau, Andonian, and Belinkov}]{meng2022locating}
Kevin Meng, David Bau, Alex Andonian, and Yonatan Belinkov. 2022.
\newblock Locating and editing factual associations in gpt.
\newblock \emph{Advances in neural information processing systems}, 35:17359--17372.

\bibitem[{Naik et~al.(2018)Naik, Ravichander, Sadeh, Rose, and Neubig}]{naik2018stress}
Aakanksha Naik, Abhilasha Ravichander, Norman Sadeh, Carolyn Rose, and Graham Neubig. 2018.
\newblock Stress test evaluation for natural language inference.
\newblock In \emph{Proceedings of the 27th International Conference on Computational Linguistics}, pages 2340--2353.

\bibitem[{Nguyen et~al.(2025)Nguyen, Huynh, Ren, Nguyen, Liew, Yin, and Nguyen}]{nguyen2025survey}
Thanh~Tam Nguyen, Thanh~Trung Huynh, Zhao Ren, Phi~Le Nguyen, Alan Wee-Chung Liew, Hongzhi Yin, and Quoc Viet~Hung Nguyen. 2025.
\newblock A survey of machine unlearning.
\newblock \emph{ACM Transactions on Intelligent Systems and Technology}, 16(5):1--46.

\bibitem[{Ouyang et~al.(2022)Ouyang, Wu, Jiang, Almeida, Wainwright, Mishkin, Zhang, Agarwal, Slama, Ray et~al.}]{ouyang2022training}
Long Ouyang, Jeffrey Wu, Xu~Jiang, Diogo Almeida, Carroll Wainwright, Pamela Mishkin, Chong Zhang, Sandhini Agarwal, Katarina Slama, Alex Ray, and 1 others. 2022.
\newblock Training language models to follow instructions with human feedback.
\newblock \emph{Advances in neural information processing systems}, 35:27730--27744.

\bibitem[{Shen et~al.(2023)Shen, Jin, Huang, Liu, Dong, Guo, Wu, Liu, and Xiong}]{shen2023large}
Tianhao Shen, Renren Jin, Yufei Huang, Chuang Liu, Weilong Dong, Zishan Guo, Xinwei Wu, Yan Liu, and Deyi Xiong. 2023.
\newblock Large language model alignment: A survey.
\newblock \emph{arXiv preprint arXiv:2309.15025}.

\bibitem[{Sucholutsky et~al.(2023)Sucholutsky, Muttenthaler, Weller, Peng, Bobu, Kim, Love, Grant, Groen, Achterberg et~al.}]{sucholutsky2023getting}
Ilia Sucholutsky, Lukas Muttenthaler, Adrian Weller, Andi Peng, Andreea Bobu, Been Kim, Bradley~C Love, Erin Grant, Iris Groen, Jascha Achterberg, and 1 others. 2023.
\newblock Getting aligned on representational alignment.
\newblock \emph{arXiv preprint arXiv:2310.13018}.

\bibitem[{Team et~al.(2024)Team, Mesnard, Hardin, Dadashi, Bhupatiraju, Pathak, Sifre, Rivi{\`e}re, Kale, Love et~al.}]{team2024gemma}
Gemma Team, Thomas Mesnard, Cassidy Hardin, Robert Dadashi, Surya Bhupatiraju, Shreya Pathak, Laurent Sifre, Morgane Rivi{\`e}re, Mihir~Sanjay Kale, Juliette Love, and 1 others. 2024.
\newblock Gemma: Open models based on gemini research and technology.
\newblock \emph{arXiv preprint arXiv:2403.08295}.

\bibitem[{Turner et~al.(2023)Turner, Thiergart, Leech, Udell, Vazquez, Mini, and MacDiarmid}]{turner2023steering}
Alexander~Matt Turner, Lisa Thiergart, Gavin Leech, David Udell, Juan~J Vazquez, Ulisse Mini, and Monte MacDiarmid. 2023.
\newblock Steering language models with activation engineering.
\newblock \emph{arXiv preprint arXiv:2308.10248}.

\bibitem[{Vaswani et~al.(2017)Vaswani, Shazeer, Parmar, Uszkoreit, Jones, Gomez, Kaiser, and Polosukhin}]{vaswani2017attention}
Ashish Vaswani, Noam Shazeer, Niki Parmar, Jakob Uszkoreit, Llion Jones, Aidan~N Gomez, {\L}ukasz Kaiser, and Illia Polosukhin. 2017.
\newblock Attention is all you need.
\newblock \emph{Advances in neural information processing systems}, 30.

\bibitem[{Wegner et~al.(1987)Wegner, Schneider, Carter, and White}]{wegner1987paradoxical}
Daniel~M Wegner, David~J Schneider, Samuel~R Carter, and Teri~L White. 1987.
\newblock Paradoxical effects of thought suppression.
\newblock \emph{Journal of personality and social psychology}, 53(1):5.

\bibitem[{Wei et~al.(2023)Wei, Haghtalab, and Steinhardt}]{wei2023jailbroken}
Alexander Wei, Nika Haghtalab, and Jacob Steinhardt. 2023.
\newblock Jailbroken: How does {LLM} safety training fail?
\newblock \emph{Advances in neural information processing systems}, 36:80079--80110.

\bibitem[{Xie et~al.(2025)Xie, Liu, and Zhang}]{xie2025erasing}
Yiwei Xie, Ping Liu, and Zheng Zhang. 2025.
\newblock Erasing concepts, steering generations: A comprehensive survey of concept suppression.
\newblock \emph{arXiv preprint arXiv:2505.19398}.

\bibitem[{Zou et~al.(2023)Zou, Wang, Carlini, Nasr, Kolter, and Fredrikson}]{zou2023universal}
Andy Zou, Zifan Wang, Nicholas Carlini, Milad Nasr, J~Zico Kolter, and Matt Fredrikson. 2023.
\newblock Universal and transferable adversarial attacks on aligned language models.
\newblock \emph{arXiv preprint arXiv:2307.15043}.

\end{thebibliography}
\end{document}